# Real Evaluations Tractability using Continuous Goal-Directed Actions in Smart City Applications

Raul Fernandez-Fernandez *, Juan G. Victores, David Estevez and Carlos Balaguer

Robotics Lab Research Group within the Department of Systems Engineering and Automation, Universidad Carlos III de Madrid (UC3M), Getafe, 28903 Madrid, Spain; jcgvicto@ing.uc3m.es (J.G.V.); destevez@ing.uc3m.es (D.E.); balaguer@ing.uc3m.es (C.B.)
* Correspondence: rauferna@ing.uc3m.es

**Abstract:** One of the most important challenges of Smart City Applications is to adapt the system to interact with non-expert users. Robot imitation frameworks aim to simplify and reduce times of robot programming by allowing users to program directly through action demonstrations. In classical robot imitation frameworks, actions are modelled using joint or Cartesian space trajectories. They accurately describe actions where geometrical characteristics are relevant, such as fixed trajectories from one pose to another. Other features, such as visual ones, are not always well represented with these pure geometrical approaches. Continuous Goal-Directed Actions (CGDA) is an alternative to these conventional methods, as it encodes actions as changes of any selected feature that can be extracted from the environment. As a consequence of this, the robot joint trajectories for execution must be fully computed to comply with this feature-agnostic encoding. This is achieved using Evolutionary Algorithms (EA), which usually requires too many evaluations to perform this evolution step in the actual robot. The current strategies involve performing evaluations in a simulated environment, transferring only the final joint trajectory to the actual robot. Smart City applications involve working in highly dynamic and complex environments, where having a precise model is not always achievable. Our goal is to study the tractability of performing these evaluations directly in a real-world scenario. Two different approaches to reduce the number of evaluations using EA, are proposed and compared. In the first approach, Particle Swarm Optimization (PSO)-based methods have been studied and compared within the CGDA framework: naïve PSO, Fitness Inheritance PSO (FI-PSO), and Adaptive Fuzzy Fitness Granulation with PSO (AFFG-PSO). The second approach studied the introduction of geometrical and velocity constraints within the CGDA framework. The effects of both approaches were analyzed and compared in the "wax" and "paint" actions, two CGDA commonly studied use cases. Results from this paper depict an important reduction in the number of required evaluations.

## 1. Introduction

In robot imitation, the user performs real-world demonstrations that are used by the robot to learn actions. One critical decision, in any imitation framework, is the selection of the model used to internally define a generalized action extracted from the demonstrations. In Programming by Demonstration (PbD), Hidden Markov Models [1] and Gaussian Mixture Models [2], are used as robot Cartesian and Joint space trajectories to represent actions. In Dynamic Motion Primitives (DMP) [3], these actions are discretized using a set of predefined control laws that achieve different Cartesian space trajectories. Finally, in Continuous Goal-Directed Actions (CGDA) actions are represented with the effects they produce on the environment, rather than being limited to the demonstrator actions that lead to them. Here, time series of scalar features extracted from the object and the environment are used to encode actions [4]. For instance, the X, Y, and Z Cartesian position of an object's centroid (three scalar features) can be used to encode an action that involves moving an object. However, an action that involves painting a wall while following no specific path can be encoded with a single scalar feature, such as the percentage of the wall that has been painted. The selection of the relevant scalar features for a specific action can be performed in two different ways: hand-crafted for the specific action or using a demonstration and feature selection algorithm [5].

One critical advantage of CGDA is that, since the encoding of actions is decoupled from the user and the robot kinematics, kinematic differences between the robot and the user do not affect the definition and transferring of actions. The CGDA framework is therefore not affected by the correspondence problem [6] presented in PbD systems.

Since robot joint trajectories are not explicitly encoded in CGDA, the framework oversees computing these trajectories. Robot joint trajectories are obtained using Evolutionary Algorithms (EA) requiring large number of evaluations. These evaluations are extremely time consuming, making them infeasible to be evaluated on the actual physical robotic platform. As a consequence of this, simulated environments are used for the execution of these evaluations, and the optimal trajectory is executed in the real robot. Smart City applications are developed in highly dynamic and complex environments, making these simulated environments models imprecise and costly to develop. The long-term goal of this work is to study the implementation of a framework that does not require a simulated environment.

An introduction to the CGDA framework is presented in Section 2. A study of the state-of-the-art methods for Evolutionary Approximation and their implementation within the CGDA framework is presented in Sections 3 and 4. Then, an introduction to Constrained Genetics Algorithms (CGA), followed by its implementation in the CGDA framework, is presented in Sections 5 and 6. The setup of the experiments performed is presented in Section 7, followed by the results obtained for each of the approaches proposed in Sections 8 and 9. To finish, a final section of conclusions is presented.

## 2. The Continuous Goal-Directed Actions Framework

Continuous Goal-Directed Actions is a framework for robot imitation in which actions are encoded as time series of object and environment features extracted from action demonstrations [4]. In Figure 1, a simplified block diagram is depicted. The CGDA framework can be used for Generalization, Recognition and Execution of actions.

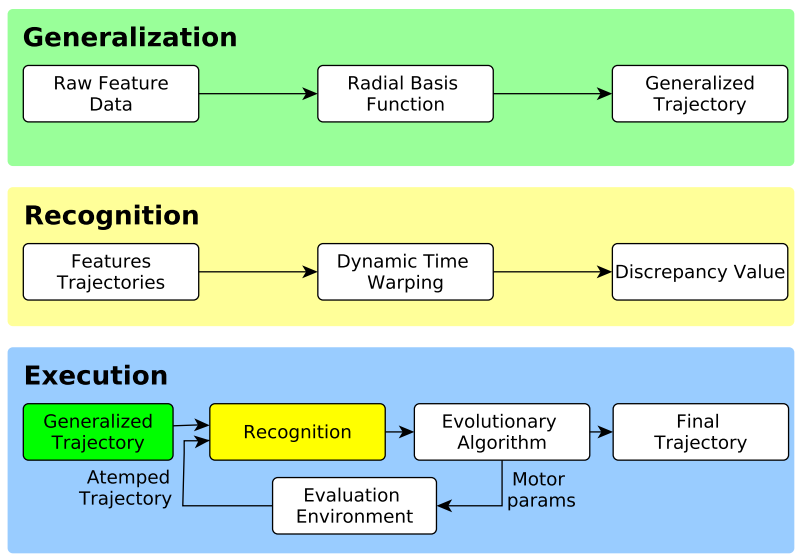

**Figure 1.** Continuous Goal-Directed Actions (CGDA) framework diagram.

The Generalization process can be seen as the encoding step, where the raw information from the demonstrations is transformed into a model of an action inside the CGDA framework. The Recognition process fulfills a double role, as it can be used to recognize similar actions, or to measure the execution performance of an action. Finally, the Execution process includes generating robot joint trajectories for an encoded action and performance in the robot environment.

*2.1. Generalization*

In the generalization process the user demonstrations are used to extract a generalized *m*-dimensional feature trajectory of the action. Here, *m* equals the number of tracked scalar features (e.g., a Cartesian coordinate, a percentage of painted wall...). First, each single action demonstration is normalized and discretized in the same time scale. Then, the generalized action is divided in *n* intermediate goals, computed as in Equation (1).

$$n = \lfloor \frac{D_{time}}{T_{min}} \rfloor. \tag{1}$$

where $D_{time}$ is the average duration of the user demonstrations, and $T_{min}$ is the minimum time interval between intermediate goals. Fixing a certain $T_{min}$, longer actions will have more intermediate goals. The generalized representation of an action $X$ is a trajectory in the *m*-dimensional feature space with *n* intermediate goals $X_j$ as defined in Matrix (2).

$$X = (X_1 \cdots X_j \cdots X_n) = \begin{pmatrix} x_{11} & \cdots & x_{1n} \\ \vdots & \vdots & \vdots \\ x_{m1} & \cdots & x_{mn} \end{pmatrix}. \tag{2}$$

Further interpolation between intermediate goals within the generalized feature trajectory can be obtained using a Radial Basis Function [7], as the one presented in Equation (3).

$$f(x) = \sum_{i=1}^{m} w_i \, \phi(\|x - x_i\|). \tag{3}$$

This interpolation is performed within the set of intermediate goals of fixed duration. A generalization example of the "wax" action is depicted in Figure 2 extracted from reference [8], where the scalar features are Cartesian coordinates that encode a geometrical action.

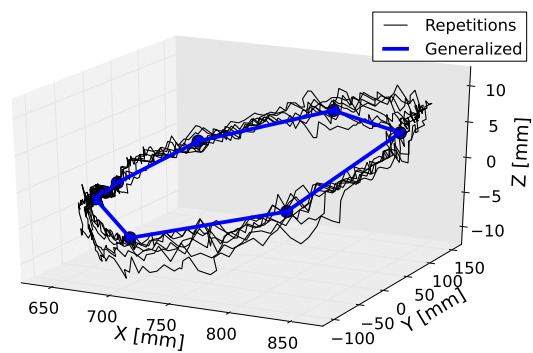

**Figure 2.** Plot of a three-feature action generalization. Black lines depict the demonstrations. The blue line is the generalized action obtained from all the demonstrations.

*2.2. Recognition*

In the recognition process, the representative generalized trajectories are used to compare the similarity between actions. This comparison is done by measuring the discrepancy between an observed action *O* and a generalized trajectory *X*. This discrepancy value measures how different are these trajectories.

Since the first implementations of CGDA, an intermediate result of the Dynamic Time Warping (DTW) algorithm is used as the discrepancy value. DTW is a technique used to optimally align two time series [9]. First, all paired combinations of points between the sequences are evaluated using a $L^2$ norm. Then, a cost matrix (4) is obtained between these references.

$$CM = \begin{pmatrix} d(o_1, x_1) & \cdots & d(o_{n'}, x_1) \\ \vdots & \vdots & \vdots \\ d(o_1, x_n) & \cdots & d(o_{n'}, x_n) \end{pmatrix}. \tag{4}$$

The lowest cost path in the cost matrix is defined as the alignment that minimize the discrepancy between these sequences. The cost $C_P(X, Y)$ of a path $P$ is calculated as the sum of the local costs as in Equation (5).

$$C_P(X, Y) = \sum_{l=1}^{L} C(o_{n'l}, x_{nl}). \tag{5}$$

where *L* is the length of the path across the *CM* matrix.

In the case of *m*-dimensional trajectories, this cost is defined as the sum of the costs of the optimal paths of each dimension, obtaining a single score *D* (6).

$$D = \sum_{i=1}^{m} C_{P_{norm}}(O_i, X_i). \tag{6}$$

This value *D* is the intermediate result used as the discrepancy value used inside the CGDA framework.

*2.3. Execution*

Joint motor parameters are not explicitly encoded in how CGDA models actions, and therefore must be generated for robot control. A conventional method such as inverse kinematics could be an option for geometrical features, in cases where the scalar features match precisely and exclusively the Cartesian coordinates of the end-effector. However, for the general case of not only geometrical features, e.g., the percentage of object or environment that has been painted, this is a not valid solution. To deal with this, several execution strategies based on Evolutionary Algorithms (EA) have been previously studied by the authors [8]. These strategies search for the optimal execution trajectory in the joint motor parameter space of the robot, using the recognition discrepancy described above as the fitness for evaluation.

From all these strategies, Incrementally Evolved Trajectories (IET) has been the most successful. The pseudocode for the IET algorithm is described in Algorithm 1. The idea behind IET is to use an evolutionary algorithm to determine the best motor joint positions $U_j$ to achieve an intermediate goal $X_j$. This evolutionary algorithm is executed in simulation.

In order to update the estimated simulator state, IET executes the previously computed motor joint position trajectory.

Although IET is general enough to be compatible with any EA, a Steady-State Tournament (SST) is classically used for the evolution step, achieving benchmarks within orders of magnitude of hundreds or thousands of evaluations.

Two different approaches are proposed to reduce the number of evaluations needed in the execution step. The first is based on reducing the number of evaluations by introducing different levels of approximations in EA. The second consists on introducing geometrical and velocity constraints to speed up the convergence.

---

**Algorithm 1** Incrementally Evolved Trajectories (IET)

    **procedure** IET($X$)
        *individuals* ←initialize
        **for** $j < n$ **do**
            **while not** *termination_conditions* **do**
                **for each** *individual* **do**
                      mental_execution($U_{[0,j-1]}$)
                      $U_j$ ←evolve($DoF$)
                      $O_j$ ←mental_execution($U_j$)
                      $D$ ←mental_recognition($O_j$, $X_j$)
                **end for**
            **end while**
        **end for**
        motor_execution($U$)
    **end procedure**

---

## 3. Introducing Approximations in Evolutionary Algorithm

This approach aims at reducing the number of evaluations in EA by introducing approximations that simplify the evolution. There are three different groups of approximation methods, classified according to what they try to approximate [10]:

- Problem Approximation: In problem approximation, a simplified version of the problem is used to replace the more complex original definition. For example, in robot navigation, a common approach is to simplify the environment representation by decomposing the environment into cells [11].
- Functional Approximation: Here, a mathematical function of reduced complexity is used to approximate the cost function of the problem. Vincenzi et al. [12] studied the effect of performing a second-order polynomial approximation of the cost function, in a Differential Evolution algorithm. The result was an improvement in the speed of the algorithm accompanied by a reduction in the number of needed evaluations.
- Evolutionary Approximation: In evolutionary approximation, the EA definition is modified to reduce the number of required evaluations. Two different kinds of methods can be found in this group: Fitness Inheritance (FI) [13] and Fitness Approximation (FA) [10].

In Fitness Inheritance, a random portion of the population is not evaluated; instead, an approximation fitness function is used to compute the fitness value of that particles. The number of required evaluations is therefore reduced by the size of that portion. In [14], FI is used to optimally select chemotherapy dose schedules. This is a complex problem (multiple drugs, schedules, effects...) where time is a critical factor.

Fitness Approximation methods are based on the idea of generating clusters in the cost function. When a new particle is generated, the particle is assigned with the fitness value of the cluster defined in the particle region without the need for evaluation. If a particle is generated in a region where there is no cluster, the particle is evaluated, and a new cluster is generated around that particle. The fitness value of this new cluster is the fitness value obtained from that particle evaluation. FA was used in [15] to reduce the number of evaluations required in bot evolution, for the computer game Unreal Tournament 2004$^{TM}$. In this scenario, evaluations are performed with a simulation at playtime, so each evaluation requires many computational resources. Another example is Bertini et al. [16], where the authors used FA to reduce another computationally expensive problem, the optimization of the start-up stage of a combined cycle power plant.

## 4. Evolutionary Approximation in Continuous Goal-Directed Actions

From all the strategies presented in the evolutionary approximation literature, Particle Swarm Optimization (PSO) is the one that has been more successful at introducing evolutionary approximations. Algorithms that introduce evolutionary approximations in PSO [17,18] have proven to converge when facing problems with non-convex functions. This is a critical advantage, since previously, a binary Genetic Algorithms (GA) with evolutionary approximations had failed to converge in this same situation [19].

In this paper, PSO and two different approximation strategies based on PSO (Adaptive Fuzzy Fitness Granulation PSO and Fitness Inheritance PSO) are introduced. These results are compared with the ones obtained using a Steady-State Tournament (SST) algorithm. SST has been the canon algorithm used for action execution since the initial proposal of the CGDA framework.

*4.1. Particle Swarm Optimization*

Contrary to other evolutionary algorithms, such as SST, where individuals are tested against each other, PSO methods are algorithms inspired on social iterations [20]. In PSO, a particle population is generated and evaluated within the search space of the problem. Then, before a new evaluation is performed, each particle is moved a certain value. This value is a function of the particle's history and some particles of the population. A typical approach is to take the particle inertia, the position of the best particle of the swarm, and the best own position of the particle.

*4.2. Adaptive Fuzzy Fitness Granulation Particle Swarm Optimization*

Adaptive Fuzzy Fitness Granulation (AFFG) is a Function Approximation (FA) method initially proposed in [21]. Similarly to other FA methods, the idea is to group the different individuals of the EA in clusters. The difference with respect other FA strategies is that in AFFG these clusters correspond to Gaussian distributions. In the experiments performed by Akbarzadeh et al. [22], AFFG achieved a reduction in the number of evaluations required of almost 90% in certain problems, without a significant increase in the performance error. The integration of AFFG with PSO, AFFG-PSO, is proposed and studied as an original contribution method of this work.

*4.3. Fitness Inheritance Particle Swarm Optimization*

Evolutionary algorithms are usually computationally expensive methods because they need to perform an evaluation step over each individual of the population. In FI [13], evaluations are performed only over a portion of the population. An approximation of the parent's fitness is used to set the fitness value of the rest of the particles.

Ducheyne et al. [19] performed a study about the feasibility of these methods in real-world scenarios. In the results the system did not converge to the global minimum when dealing with non-convex problems. However, later, Reyes et al. [17] performed the study of the integration of FI in PSO algorithms (FI-PSO). The results obtained from this study were quite different. FI-PSO can reach an optimal solution even when dealing with non-convex problems. The same authors proposed different modifications of the initial algorithm in reference [18]. Here, the implementation of a FI-PSO algorithm, using a flight formula for the fitness approximation similar to the one used in PSO, obtained the best results.

## 5. Constrained Genetic Algorithms

The second studied approach to reduce the number of evaluations using EA consisted on introducing constraints in GA [23]. The group of GA is the group of EA that includes SST. Constrained GA have been introduced as a modification of the initial idea of GA. Here, constraints are introduced in the algorithm with the goal to simplify the problem, or to model solutions that in the real world are not feasible for the system. Constrained Genetic Algorithms can be classified as a function of how they deal with solutions outside the valid space (infeasible solutions) in different strategies [24,25]:

- Rejecting strategy: In Rejecting strategy, only the feasible solutions are used by the algorithm, while infeasible solutions are straightforwardly rejected. This strategy fails if all the solutions in a given iteration are infeasible. Yong and Sannomiya [26] applied this strategy to a flowshop problem, where many solution combinations are not feasible ones. Here, constraints were introduced to reduce the search space to only valid combinations.
- Repairing strategy: Contrary to the Rejecting Strategy, this strategy tries to convert all the infeasible solutions into feasible ones. A repairing step is therefore introduced to achieve this. However, this repairing step can be sometimes as complex as the problem itself. Chootinan and Chen [27] defined a gradient function in the constrained space that later was used for the repairing step. Infeasible solutions were redirected to the valid space using this gradient.
- Modifying genetic operator strategy: In this strategy, the hyperparameters of the algorithm are tuned to only produce feasible solutions in the GA. This is the case, for example, of introducing constraints in the EA individuals.
- Penalty strategy: In the penalty strategy, a fitness penalty parameter is introduced to penalize infeasible solutions. The constrained space is converted into an unconstrained one with penalty regions. These penalties parameters can be defined in two different ways:

$$eval(x) = \begin{cases} f(x), & if \quad x \in F \\ f(x) + p(x) & otherwise \end{cases}. \tag{7}$$

$$eval(x) = \begin{cases} f(x), & if \quad x \in F \\ f(x)p(x) & otherwise \end{cases}. \tag{8}$$

Here, $f(x)$ is the cost function, $p(x)$ defines the penalty function, and $F$ is the feasible space. The additive option (7) is the most used from the two presented.

From all the options presented, the penalty strategy is the most used one in the Constrained GA literature. Different penalty strategies can be used to compute the penalty value [28]. Death penalty strategies assign a $\infty$ fitness penalty value to the infeasible solution. In Static penalties, a finite constant value is used. In Dynamic penalties, the penalty is set as a function of the number of generations. Adaptive penalties use the state of the population to change the penalty value. An example of Adaptive penalties is a Constrained GA where, if all the solutions obtained in the last iteration are infeasible, the penalty value $p(x)$ is increased, on the contrary, if all are feasible is decreased [29]. Finally, in Annealing Penalties, Annealing Algorithms are used to calculate the penalty value using an internal temperature parameter [30]. In addition to this, several authors have proposed different modifications of these base methods to deal with Constrained GA [31–34].

Some other works use combinations of these penalty strategies to define the constrained problem. Chang et al. used constrained GA to control the water release of a water reservoir as an optimization problem [35]. Here, two different penalty strategies were used. A Death penalty was introduced to penalize invalid solutions where the water level was outside the reservoir boundaries. At the same time, Static penalties was introduced to penalize solutions outside the safe water storage levels for each season.

**6. Constrained Genetic Algorithms in Continuous Goal-Directed Actions**

Since the goal of CGDA is to work with generalized actions, the goal, here, is to define a set of constraints that do not limit the generalized nature of CGDA. To achieve this, two non-action-specific constraints are integrated within the CGDA framework. These two constraints are the geometrical and velocity constraints.

The geometrical constraint defines a valid Cartesian space inside the search function. Solutions outside this valid space are considered infeasible for the system. For the definition of this space, any geometry, such as a bounding box, a bounding sphere, a convex hull, or any polyhedron, can be defined to fit the solution space of the action. To reduce computational times, in this paper, this valid space is given as the axis-aligned minimum bounding box, with a fixed dilatation value on all the axes, around the solution space.

Since the goal is continuously changing in CGDA, controlling the exploration behavior of the particles is a key factor to speed up convergence. This exploration rate can be controlled by controlling the velocity of the particles inside the search space. High velocity value constraints favor exploration, while low velocity value favor exploitation. The velocity constraint is defined as an internal hyperparameter of the GA. Each iteration, the velocity value of each particle in the joint space is obtained and compared with a threshold. If this value is bigger, then, this new particle is considered outside the valid space and penalized.

For both approaches a Penalty Strategy is used to deal with solutions outside the valid space. The Death Penalty strategy was the chosen Penalty Strategy for the introduction of penalties, due to the amount of literature related.

## 7. Experiments

Both approaches were tested in the experiments with the goal to measure and compare their impact in the CGDA framework. The number of evaluations required for the evolution step was the main parameter used to measure this performance. Two different sets of experiments were performed in this work, one for each presented approach. In the first set of experiments, Evolutionary Approximation was introduced in the CGDA framework. Here, four different evolutionary algorithms were used: Steady-State Tournament (SST) [36] (used in the original proposal of CGDA), naïve PSO [20], AFFG-PSO (this is an original contribution of a modified version of PSO with Adaptive Fuzzy Fitness Granulation [22]), and the Fitness Inheritance PSO (FI-PSO) algorithm as proposed in [18]. In the second sets of experiments, geometrical and velocity constraints were introduced in the CGDA framework. Here, the SST algorithm was used in the evolution step to establish a common baseline, as SST was the canon algorithm used since the initial proposition of CGDA.

Previous works using SST within the CGDA framework required large numbers of evaluations to converge. With this in mind, evaluations were computed using a simulation environment inside OpenRAVE [37]. TEO (Model available at https://github.com/roboticslab-uc3m/teo-main), the humanoid robot from the Robotics Lab of Universidad Carlos III de Madrid [38], was the robotic platform used both for simulation and real action execution. Three of the six Degrees of Freedom of the right arm of the robot were used for the experiments, all other joints were set to a fixed position. These methods and constraints were integrated and open-sourced (Source code available at https://github.com/roboticslab-uc3m/xgnitive) within the CGDA architecture. For all the experiments, the execution of the "wax" (known as "clean" in previous literature) and the "paint" actions [8] were performed, choosing IET as the evolutionary strategy. The "wax" action is a representative example of a geometrical action, with the particularity that intermediate positions are also relevant for action execution. This is an example where classical geometrical imitation frameworks tend to shine. The "paint" action is an example of an action where classical geometrical imitation frameworks fail due to the disparity in the demonstrations. The goal using these actions is double. First, we demonstrate that CGDA is able two work in both scenarios. Second, we use the common baseline to compare the results with previous works of CGDA [8].

In a "wax" action, the centroid of a grasped object (e.g., a sponge for waxing) is moved in a circular trajectory, for one revolution, of 30 cm of diameter. Three scalar features (X,Y,Z), corresponding to the position of the object's centroid in the Cartesian space, are used to define the action. This setting was chosen with the intention to be similar to solve as an inverse kinematics problem. This was done to demonstrate how the kinematics problem is solved within the expected ranges of the demonstration, despite the system being agnostic to the nature of the tracked features. The goal of the "paint" action is to paint a wall. To emulate the process, if the painting tool is closer than a fixed "painting distance" to the wall, that part of the wall is considered painted. In this action only one scalar feature, the percentage of painted wall, is used. In Figure 3 there is an example of this action execution.

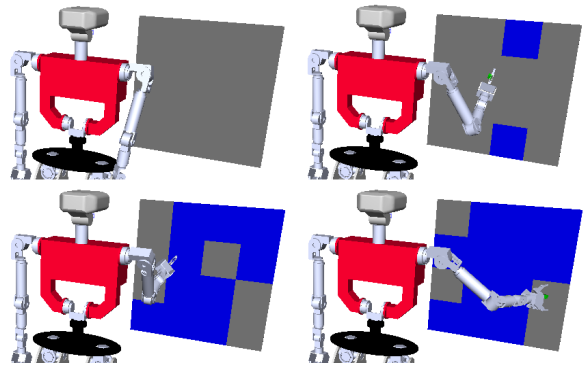

**Figure 3.** Execution of the "paint" action using the humanoid robot TEO inside a simulated environment.

In terms of EA setup, the movements of the joints of the humanoid were restricted between $-15$ and $100$ degrees. The individual mutation probability for the SST algorithm was set to 60%. In the "wax" action, a population of 50 individuals (three joint parameters) was used for the evolution. A solution was considered found when reached 3 consecutive generations without improvement. In the "paint" action the population of individuals was set to 10. The termination condition increased to 10 consecutive generations without improvement or achieving a zero error in the feature trajectory. This increase was due to the computational cost differences between these two actions. For the "wax" action, a higher number of required evaluations was expected than in the case of the "paint" actions.

Two different parameters were measured during the experiments, the number of required evaluations, and the discrepancy value. While the main goal of this work is to reduce the number of evaluations, the discrepancy value was measured to study the impact of this reductions in the performance of the action. For the "paint" action, the percentage of painted wall at the end of the action was also obtained as an additional more intuitive measure of the action performance.

## 8. Evolutionary Approximation Results

Some additional tuning was required for the additional methods proposed in the Evolutionary Approximation Experiments. For each PSO algorithm, a PSO inertia weight value of 1.2 was used, fixing the maximum particle velocity to 5. In the case of AFFG-PSO, a maximum number of 3 simultaneous granules was set in the cost function [22]. Finally, a 55% of inheritance proportion was used for the FI-PSO method [18].

*8.1. Wax*

The average results after running 50 repetitions of the "wax" action are depicted in Table 1. In Figure 4, the cumulative value of required evaluations for each of the methods is represented as a function of the intermediate goal.

These results were compared with the ones obtained with the original SST algorithm. FI-PSO was the method that obtained the best results in terms of reducing the number of evaluations. The number of evaluations was reduced a 65% with this method, with a 25% increase in the discrepancy.

In Figure 5, a comparison between the executed trajectory and the generalized one is represented for each of the proposed methods.

**Table 1.** "Wax" Evolutionary Approximation results.

| Method | Evaluations | Discrepancy |
| --- | --- | --- |
| SST | 9679 | 274 |
| PSO | 8470 | 213 |
| AFFG-PSO | 5314 | 434 |
| FI-PSO | 3432 | 362 |

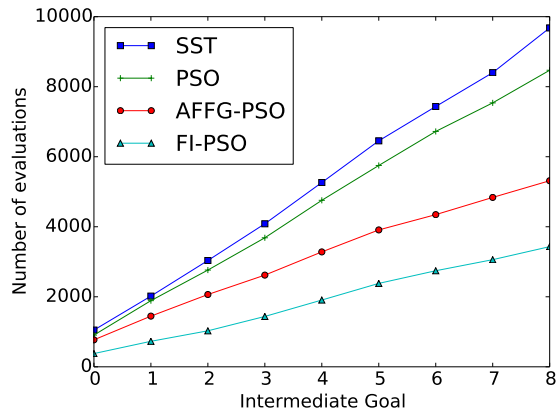

**Figure 4.** "Wax" Evolutionary Approximation results: Number of cumulative evaluations for each intermediate goal.

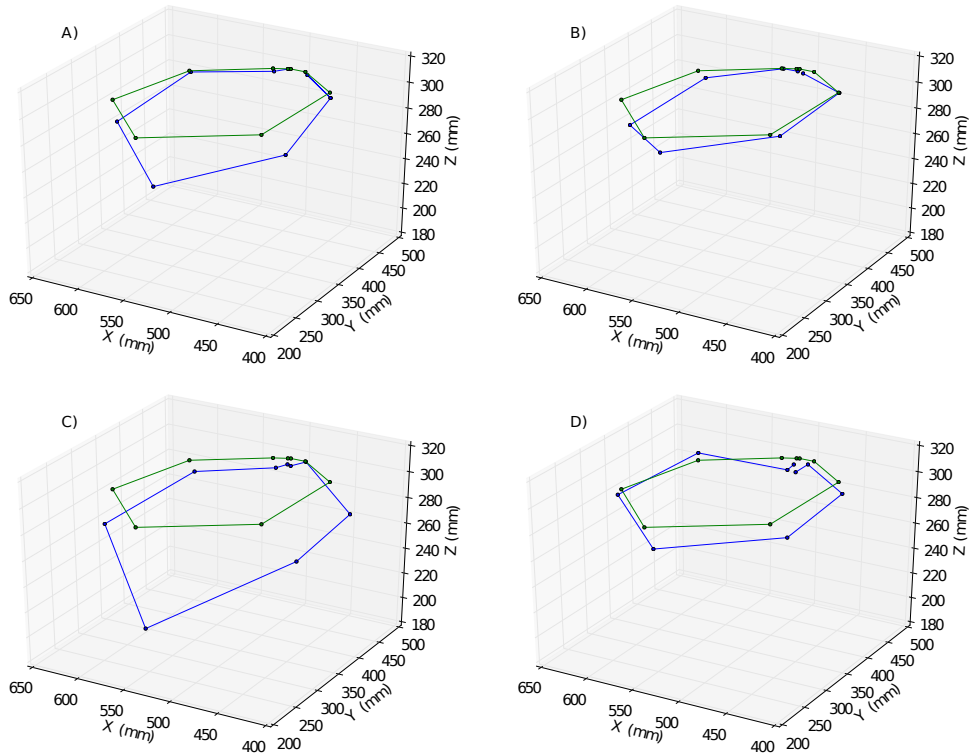

**Figure 5.** "Wax" Evolutionary Approximation results: The generalized trajectory is represented with the green trajectory, the blue trajectory is the executed one. The methods corresponding to each of the cases are the following: (**A**) SST; (**B**) PSO; (**C**) AFFG-PSO and (**D**) FI-PSO.

*8.2. Paint*

The average results obtained after 100 executions of the "paint" action are represented in Table 2. In Figure 6 there is a representation of the number of cumulative evaluations required at each intermediate goal, with a representation of the generalized trajectory and all the obtained trajectories.

Table 2. "Paint" Evolutionary Approximation results.

| Method | Evaluations | Discrepancy | Painted (%) |
|---|---|---|---|
| SST | 539 | 7.25 | 94.4 |
| PSO | 583 | 12.06 | 91.44 |
| AFFG-PSO | 537 | 16.56 | 89.75 |
| FI-PSO | 441 | 20.13 | 87.88 |

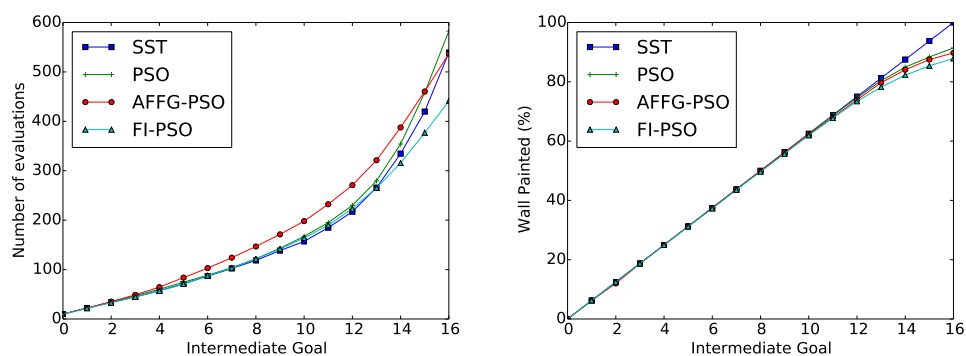

**Figure 6.** "Paint" Evolutionary Approximation results: Number of cumulative evaluations at each intermediate goal (**Left**). Comparison between the generalized trajectory and the executed ones (**Right**).

FI-PSO was the best method in terms of evaluation reduction. Compared to the original SST algorithm, FI-PSO required 17% less evaluations while having a tradeoff of reaching a 7% reduction of painted wall.

## 9. Constrained Genetic Algorithms Results

In this set of experiments, aks SST method, with the same hyperparameters as the ones used in the Evolutionary Approximation experiments, was used. Different constraint values were introduced to both actions to study the impact of using geometrical and velocities constraints inside the CGDA framework.

*9.1. Wax*

In the case of the "wax" action, six different dilatation values (distance between the constraints and the solution space) were used: 0.01 m, 0.05 m, 0.1 m, 0.2 m, 0.3 m and ∞. Here, a ∞ dilatation value is equivalent to using the original SST algorithm. Table 3 depicts the average results after running 50 repetitions of the "wax" action.

Table 3. "Wax" geometrical constraint experiment results.

| Dilatation (m) | 0.01 | 0.05 | 0.1 | 0.2 | 0.3 | ∞ |
|---|---|---|---|---|---|---|
| **Evaluations** | 3212 | 3163 | 2993 | 4960 | 5722 | 9679 |
| **Discrepancy** | 465 (2) | 503 (1) | 471 (3) | 312 | 331 | 274 |

Here, in the first three cases (0.01 m, 0.05 m, 0.1 m), the value in parenthesis () shows the number of solutions that ended outside the valid space and therefore assigned with a Discrepancy = ∞.

This discrepancy value was due to executed trajectories resulting outside the valid space, and therefore assigned with a death penalty value. These trajectories were not computed within the average of the experiments.

Figure 7 depicts the value of cumulative evaluations required at each intermediate goal. These results are again compared with respect to the original SST algorithm. The highest reductions in terms of evaluations were obtained using dilatation values $\leq$0.1 m. A reduction in the number of evaluations around the 60% was obtained using these dilatation values, meaning a reduction of over 6000 required evaluations. These dilatation values also affected the performance of the executed trajectory, obtaining the cited invalid solutions.

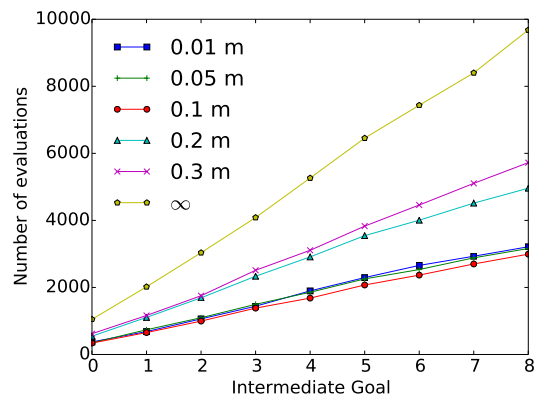

**Figure 7.** "Wax" geometrical constraint experiment results: Number of cumulative evaluations required at each intermediate goal, as a function of the dilatation value (0.01 m, 0.05 m, 0.1 m, 0.2 m, 0.3 m and $\infty$).

In the second set of constraints used in the "wax" action, the velocity constraint was introduced in the CGDA framework. The average results of these experiments after 50 repetitions of the "wax" action, using different velocity constraints values, are represented in Table 4. In Figure 8, the cumulative number of evaluations required by each of the methods at each intermediate goal is represented.

**Table 4.** "Wax" velocity constraint experiment results.

| Max. Velocity | 5 | 10 | 20 | 60 | 80 | $\infty$ |
|---|---|---|---|---|---|---|
| **Evaluations** | 3591 | 4058 | 5723 | 6876 | 7349 | 9679 |
| **Discrepancy** | 540 | 483 | 331 | 346 | 330 | 274 |

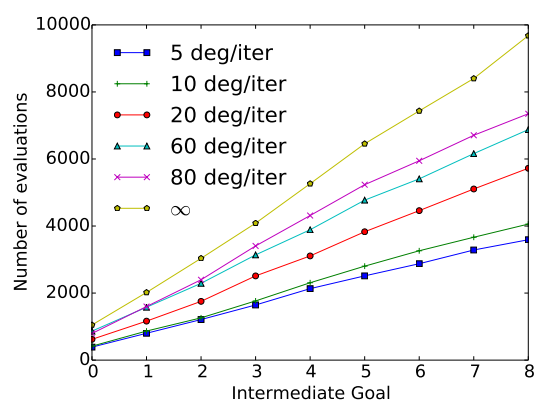

**Figure 8.** "Wax" velocity constraint experiment results: Number of cumulative evaluations required at each intermediate goal as a function of the velocity constraint (5, 10, 20, 60, 80 and $\infty$ degrees/iteration).

These results show how a reduction in the velocity threshold comes with a reduction in the number of evaluations. This reduction reaches a maximum value when using a 5 degree/iteration constraint, needing 63% less evaluations with respect to the original SST method (Corresponding with a ∞ velocity constraint). This reduction, however, comes with a 97% increment in the discrepancy value.

*9.2. Paint*

In the case of the "paint" action, the same dilatation values were used for the geometrical constraint experiments. The average results obtained after 100 repetitions of the "paint" action are depicted in Table 5. In Figure 9 the cumulative number of evaluations required at each intermediate goal is represented.

**Table 5.** "Paint" geometrical constraint experiment results.

| Dilatation (m) | 0.01 | 0.05 | 0.1 | 0.2 | 0.3 | ∞ |
|---|---|---|---|---|---|---|
| **Evaluations** | 319 | 307 | 220 | 334 | 360 | 539 |
| **Discrepancy** | 193 | 17 | 5.7 | 7.3 | 6.3 | 7.3 |
| **Painted Wall (%)** | 64.06 | 89.58 | 95.13 | 94.38 | 94.56 | 94.44 |

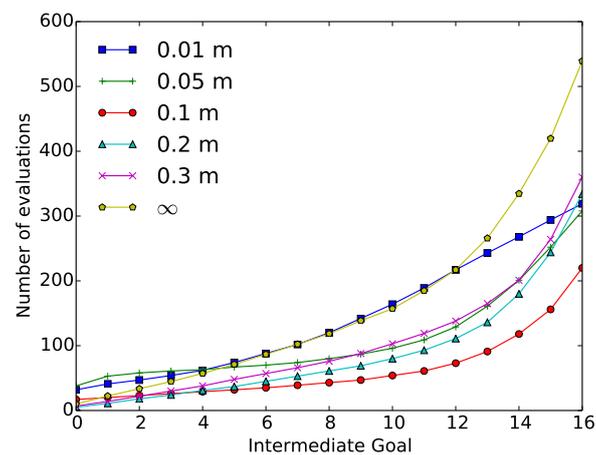

**Figure 9.** "Paint" geometrical constraint experiment results: Number of cumulative evaluations required at each intermediate goal as a function of the velocity constraint (0.01 m, 0.05 m, 0.1 m, 0.2 m, 0.3 m and ∞).

These results were again compared with the original SST algorithm (which corresponds to ∞ velocity constraint). In this scenario, the 0.1 m constraint is the one with the best performance. Using this constraint, a reduction in the number of required evaluations about 60% was obtained while having a 22% lower discrepancy. As the dilatation value moves away from this point, both the number of required evaluations, and the discrepancy value are increased. The average results obtained with the velocity constraint, in terms of required evaluations and discrepancy, are represented in Table 6. Here, six different velocity constraints were used (20, 60, 80, 100, 120 and ∞ degrees/iteration). Figure 10 depicts the cumulative number of required evaluations at each intermediate goal.

**Table 6.** "Paint" velocity constraint experiment results.

| Max. Velocity | 20 | 60 | 80 | 100 | 120 | ∞ |
|---|---|---|---|---|---|---|
| **Evaluations** | 543 | 557 | 572 | 527 | 529 | 539 |
| **Discrepancy** | 24.8 | 12.8 | 10.4 | 8.1 | 8 | 7.3 |
| **Painted Wall (%)** | 87.06 | 90.21 | 92.68 | 93.75 | 93.88 | 94.44 |

The highest reduction in the number of evaluations is obtained using the 100 degree/iteration constraint, with a reduction of 2.2% compared with the original SST algorithm. The maximum number of required evaluations in the experiments is obtained using an 80 degree/iteration constraint, with an increment in the number of evaluations of 6%. All the constrained scenarios obtained a bigger discrepancy value. The more constrained the system is, the higher the discrepancy value obtained.

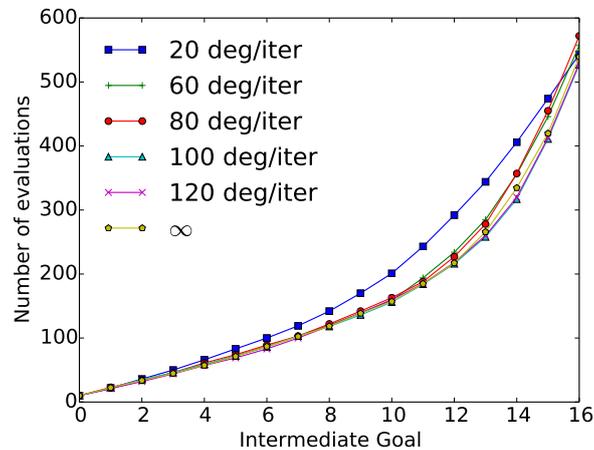

**Figure 10.** "Paint" velocity constraint experiment results: Cumulative number of evaluations at each intermediate goal for different Velocities (20, 60, 80, 100, 120 degree/iteration and ∞).

## 10. Conclusions

In this paper, Evolutionary Approximation and Constrained Genetic Algorithms, two different approaches to reduce the number of evaluations in Evolutionary Algorithms (EA), were proposed to deal with real-world Smart City applications challenges. In Evolutionary Approximation, approximations were introduced inside the EA to approximate the fitness value of a proportion of the particle population without the need for evaluations. Here, three different methods were introduced inside the CGDA framework: Particle Swarm Optimization (PSO), Fitness Inheritance PSO (FI-PSO) and Adaptive Fuzzy Fitness Granulation PSO (AFFG-PSO). Although PSO is not an Evolutionary Approximation algorithm, it was introduced to compare the results obtained by both FI-PSO and AFFG-PSO with the vanilla method. Here, FI-PSO and in a smaller degree AFFG-PSO were able to introduce critical reductions in the number of evaluations. The results from the experiments depicted reductions in the number of evaluations of 65% "wax" action and 17% for the "paint" action, compared with the original SST method. These results came at the cost of introducing a higher discrepancy in the executed trajectory. However, this discrepancy increase did not come with a significant deterioration of the executed trajectories (see Figure 5 for the "wax" action, and Figure 6 (Right) for the "paint" action). In Figure 5, the executed trajectory obtained with FI-PSO is smoother than the one obtained with SST needing 65% less iterations, although in fact it has a higher discrepancy. The hypothesis here is that this is a result of how evolutionary approximation algorithms works. These algorithms tend to behave worse when they are near the convergence area. In the case of AFFG, having large granules near this area can make the algorithm to diverge. As a result of this, stationary errors are common in these algorithms. Using IET, these stationary errors are accumulated for each intermediate goal, meaning an increase in the total discrepancy value obtained, and producing the higher discrepancy value appearing in the results. Some work has already been done to reduce this effect, in AFFG for example [22], where the granules shrink when reaching lower fitness value. However, studying this effect and how to deal with it, is outside the scope of this work and will be introduced in future works of the authors.

The second approach, Constrained Genetic Algorithms, introduced two different constraints, the velocity constraint and the spatial constraint. The results obtained with the velocity constraint did not reach the expectations of this work. In the case of the "wax" action, important reductions in the number of evaluations came with equally important increases in the discrepancy value, see Table 4 and Figure 8 for reference. For the "paint" action these increases in discrepancy did not come with an equal improvement in the number of evaluations, see Table 6 and Figure 10 for reference. One hypothesis about the differences in the results obtained using this action comes from the idea that "paint" is not a sequential action such as "wax". When executing the "paint" action, the robot may need to perform large movements (high velocity values) to paint loose squares of the wall that have not been painted in the first swept of the robot (low velocity values). Using a fixed velocity threshold is not an efficient solution for this action. In addition to this, a possible explanation about the loss of performance after introducing this constraint in both actions can be found in past works related to CGDA. In [39], the authors introduced Sequential Incremental Combinatorial Search of motor primitives for the execution of actions inside the CGDA framework. The idea was to use different sets of primitives with different levels of constraints. In the experiments, the less constrained sets (including large and short primitives) obtained better results compared to the constrained ones (including only short primitives). Here, the authors concluded that the less constrained sets had richer possibilities, allowing a better performance of the CGDA framework. In these experiments, the velocity constraint is producing a similar effect reducing the possibilities of the system. This explains the performance reduction experience with this action.

In the case of geometrical constraints, there are important differences between the results obtained with both actions. With respect to the SST algorithm, in the "wax" action the algorithm needed 60% less evaluations while obtaining a 72% increase error in the discrepancy value. The "paint" action obtained the same results in terms of evaluation reduction, but in this case also decreasing a 22% the discrepancy value compared the SST algorithm. The hypothesis behind these results is that the shape of the constrained region, in this case a box, is critical for the performance of this constraint. This shape fits perfectly the "paint" solution space (a box), but not the "wax" solution space (a circle). This hypothesis is asserted comparing the results obtained with each action. Some previous knowledge of the action is required for the correct introduction of these constraints in the EA.

**Author Contributions:** Conceptualization, R.F.-F. and J.G.V.; Data curation, R.F.-F.; Formal analysis, J.G.V.; Funding acquisition, C.B.; Investigation, R.F.-F. and J.G.V.; Methodology, R.F.-F.; Project administration, J.G.V. and C.B.; Resources, C.B.; Software, R.F.-F., J.G.V. and D.E.; Supervision, J.G.V. and C.B.; Validation, R.F.-F.; Visualization, J.G.V.; Writing—original draft, R.F.-F.; Writing—review and editing, J.G.V., D.E. and C.B.

**Funding:** The research leading to these results has received funding from the RoboCity2030-III-CM project (Robótica aplicada a la mejora de la calidad de vida de los ciudadanos. fase III; S2013/MIT-2748), funded by Programas de Actividades I+D en la Comunidad de Madrid and cofunded by Structural Funds of the EU.

**Conflicts of Interest:** The authors declare no conflict of interest.